\documentclass[10pt,twocolumn,letterpaper]{article}

\usepackage[pagenumbers]{cvpr} % To force page numbers, e.g. for an arXiv version

\usepackage{microtype}
\usepackage[table]{xcolor}

\definecolor{cvprblue}{rgb}{0.21,0.49,0.74}
\usepackage[pagebackref,breaklinks,colorlinks,allcolors=cvprblue]{hyperref}
\usepackage{multirow}

\usepackage{listings}    % For code blocks
\usepackage{xcolor}      % For syntax highlighting
\usepackage{float}       % To allow [H] placement for algorithms
\usepackage{caption}     % Better captions
\usepackage{algorithm}   % Algorithm float environment
\usepackage{adjustbox}

\lstdefinestyle{ieeepython}{
    language=Python,
    basicstyle=\ttfamily\footnotesize,
    keywordstyle=\color{blue}\bfseries,
    commentstyle=\color{green!60!black}\itshape,
    stringstyle=\color{red},
    showstringspaces=false,
    breaklines=true
}

\def\paperID{xxxxx} % *** Enter the Paper ID here
\def\confName{CVPR}
\def\confYear{2026}

\title{ConfCtrl: Enabling Precise Camera Control in Video Diffusion via Confidence-Aware Interpolation}

\author{
Liudi Yang$^{1}$,
George Eskandar$^{4}$,
Fengyi Shen$^{3}$,\\
Mohammad Altillawi$^{4}$,
Yang Bai$^{2}$,
Chi Zhang$^{4}$,
Ziyuan Liu$^{4}$\thanks{corresponding author},
Abhinav Valada$^{1}$\\[0.5em]
$^{1}$University of Freiburg,\\
$^{2}$Ludwig Maximilian University of Munich,\\
$^{3}$Technical University of Munich,\\
$^{4}$Huawei Heisenberg Research Center (Munich)\\
$^{*}$Corresponding author
}

\begin{document}

\twocolumn[{
\renewcommand\twocolumn[1][]{#1}
\maketitle
\begin{center}
    \centering
    \vspace{-15pt}
    \includegraphics[width=0.9\linewidth]{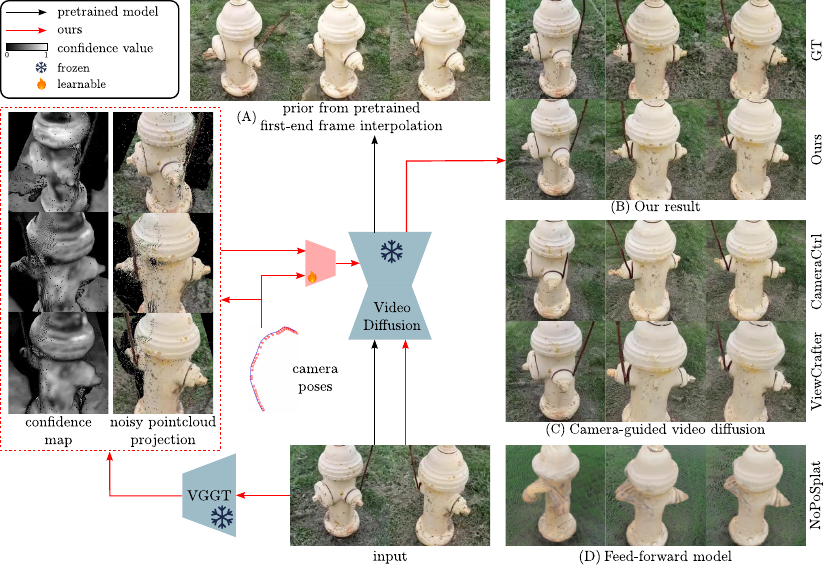}
    % \vspace{-15pt}
\captionof{figure}{Overview of ConfCtrl. (A)~demonstrates the strong prior of a first– and end–frame video interpolation model. This is fully inherited by our model for further geometric improvement.
(B)~ \textbf{Our ConfCtrl enables precise camera control in video diffusion models.}
(C)~shows that camera-guided video diffusion models struggle to strictly follow the target camera poses.
(D)~indicates that feedforward models underperform due to limited generative capacity, leading to inaccurate estimation of Gaussian Splat parameters.}
    \label{fig:teaser}
\end{center}
}]

\begin{abstract}
We address the challenge of novel view synthesis from only two input images under large viewpoint changes. Existing regression-based methods lack the capacity to reconstruct unseen regions, while camera-guided diffusion models often deviate from intended trajectories due to noisy point cloud projections or insufficient conditioning from camera poses. To address these issues, we propose \textbf{ConfCtrl}, a confidence-aware video interpolation framework that enables diffusion models to follow prescribed camera poses while completing unseen regions. ConfCtrl initializes the diffusion process by combining a confidence-weighted projected point cloud latent with noise as the conditioning input. It then applies a Kalman-inspired \emph{predict–update} mechanism, treating the projected point cloud as a noisy measurement and using learned residual corrections to balance pose-driven predictions with noisy geometric observations. This allows the model to rely on reliable projections while down-weighting uncertain regions, yielding stable, geometry-aware generation. Experiments on multiple datasets show that ConfCtrl produces geometrically consistent and visually plausible novel views, effectively reconstructing occluded regions under large viewpoint changes.
\end{abstract}    
\section{Introduction}
\label{sec:intro}

Novel view synthesis from sparse input views, especially under large viewpoint changes, remains a challenging problem that has attracted significant attention. Although recent methods~\cite{pixelsplat,noposplat,seva,uni3c,anysplat,ac3d,cameractrl,dimensionx,cat3d,cat4d,lvsm} have achieved remarkable progress, their performance often degrades when the input views are captured from widely separated viewpoints. Overcoming this limitation is crucial for real-world applications, where only a few images with large viewpoint differences are typically available.

Existing approaches can be broadly categorized into \textit{regression-based} and \textit{diffusion-based} methods. Regression-based methods learn explicit 3D scene representations in a feedforward manner and render images conditioned on target camera poses. While these methods can accurately follow the specified camera trajectory, they often lack sufficient generative capability under sparse or incomplete observations, resulting in noticeable rendering artifacts (Figure~\ref{fig:teaser}(D)). In contrast, diffusion-based approaches benefit from powerful generative priors and strong completion ability gained through large-scale pretraining. However, current camera-guided diffusion models, such as CameraCtrl~\cite{cameractrl} (Figure~\ref{fig:teaser}(C)), still struggle to precisely follow target camera poses, limiting their controllability in view synthesis.  

A potential solution is to either enhance regression-based methods with stronger generative capabilities or enable diffusion-based methods to strictly follow desired camera trajectories. We observe that the first approach requires substantial training data to achieve satisfactory performance, even when integrating pretrained video diffusion to inject generative latents into a Gaussian decoder~\cite{splatflow,wonderland}. Alternatively, the second approach demands a carefully designed camera-conditioning mechanism that ensures precise adherence to target camera motions while generating plausible scene contents with temporal and spatial consistency at the same time. Although this approach is advantageous in reusing the rich priors of pretrained video diffusion models, thereby mitigating the need for extensive training data, the main challenge is twofold: (1)~the single-image-to-video generation paradigm tends to accumulate deviation over frames, and (2)~the absence of explicit 3D priors limits accurate geometric consistency.

To address problem~(1), we observe that employing a video interpolation model can naturally enhance geometric consistency through its head–tail interpolation mechanism (Figure~\ref{fig:teaser}(A)). For problem~(2), prior works \cite{uni3c,gen3c,viewcrafter,spatialcrafter} incorporate projected point clouds obtained from monocular depth estimation models. With this strong 3D prior in pixel space, the model can achieve better camera following capability if the projected point cloud is sufficiently accurate. In practice, however, these approaches often suffer from distortion and scale ambiguity of real point cloud estimation, and conditioning on such noisy inputs often leads to suboptimal performance (ViewCrafter in Figure~\ref{fig:teaser}(C)). Recent 3D foundation models~\cite{vggt,dust3r,mast3r} offer more reliable 3D priors by estimating projected point clouds from paired images, leading to improved geometric accuracy. However, these models are still imperfect, as residual artifacts and warping can degrade the quality of the conditioning signal. This motivates a camera condition mechanism that can compensate for \textit{uncertainty} in projected point clouds. 

Building on these insights, we propose ConfCtrl, a confidence-aware framework for novel view synthesis under large viewpoint changes. ConfCtrl builds upon a pretrained video interpolation model and explicitly models the uncertainty in point clouds projected from 3D foundation models. By incorporating confidence-aware conditioning, ConfCtrl enables the video diffusion model to more precisely follow target camera poses, effectively exhibiting feedforward model-like behavior. Concretely, instead of initializing the rectified flow model from pure noise, we start from a sum of a confidence-weighted projected point cloud latent and noise. This initialization improves the adaptation from pretrained temporal interpolation knowledge toward novel view synthesis. The confidence maps act as reliability weights that quantify the trustworthiness of each projected point in the latent space.

Second, inspired by the well-known Kalman Filter~\cite{kalman}, we introduce a camera conditioning mechanism that jointly encodes projected point clouds and camera poses to mitigate uncertainty in 3D geometric priors. This mechanism operates through two submodules: a prediction submodule, which conditions the latent state on the target camera pose, and an update submodule, which uses the projected point cloud as a noisy measurement to refine the prediction, improving stability and precision in camera control. We validate the effectiveness of our approach across multiple datasets, consistently surpassing the baseline methods. Furthermore, by leveraging the generalization capabilities of pretrained video diffusion models, our method exhibits strong zero-shot performance on out-of-distribution scenarios. 

Our main contributions are summarized as follows:
\begin{itemize}[topsep=0pt]
\item We show that, under the challenging sparse inputs setting, leveraging a pretrained video interpolation model will provide stronger 3D consistency for novel view synthesis. 
\item We introduce a diffusion initialization strategy using a latent derived from the projected point cloud, enabling more effective adaptation from interpolation to novel view synthesis. 
\item We propose a camera-pose conditioning mechanism that jointly encodes projected point clouds and camera poses via a \emph{predict–update} architecture, handling uncertainty in geometric priors to achieve robust geometry and precise camera control.
%\mo{place this as the first contribution. Introducing something is stronger than showing a result (show that...)}
\item Extensive experiments on multiple datasets demonstrate that our method consistently outperforms existing baselines and achieves strong zero-shot generalization.
\end{itemize}
\section{Related Work}
\label{sec:related}

{\parskip=2pt
\noindent\textbf{Regression-based Methods}: 
With the rapid progress of 3D neural rendering~\cite{gs,nerf}, feedforward approaches have gained increasing attention due to their generality and fast inference. Recent methods~\cite{pixelsplat,noposplat,depthsplat,mvsplat,splat3r,transplat,efreesplat,instantsplat,selfsplat,spfsplat,lrm,instant3d,gslrm,longlrm} directly infer 3D scene structures from input images and regress rendering parameters for novel view synthesis. However, their performance degrades significantly with sparse inputs and large viewpoint changes. This limitation mainly arises from the lack of generative priors, which hinders accurate 3D representation and the inference of unseen regions. AnySplat~\cite{anysplat} alleviates this issue by incorporating the foundation model VGGT~\cite{vggt} for Gaussian positioning, but it still cannot hallucinate unseen regions. LVSM~\cite{lvsm} adopts a data-driven design with minimal 3D inductive biases, yet remains challenged by large viewpoint variations. To improve generative capability, later works~\cite{latentsplat,mvsplat360,3dgsenhancer,genfusion} introduce generative models as refinement modules for scene completion or Gaussian deblurring. Nevertheless, these approaches depend on noisy renderings as input, leading to performance degradation when the initial Gaussian reconstruction is severely corrupted.}

{\parskip=2pt
\noindent\textbf{Diffusion-based Methods with Camera Control}: 
Another line of research focuses on leveraging diffusion models for novel view synthesis. Recent works~\cite{motionctrl,cameractrl,ac3d,seva,trajectorycrafter,realcami2v,recamaster,causnvs,gen3r} introduce camera pose conditioning into video diffusion models, allowing them to synthesize viewpoint-varying sequences. These methods provide video diffusion models with controllable generative power but still suffer from limited camera controllability and the absence of explicit 3D priors, leading to inconsistent content generation and imperfect adherence to target camera trajectories. To mitigate this problem, several approaches~\cite{viewcrafter,uni3c,gen3c,voyager,flexworld,spatialcrafter,reconx} estimate depth from images and use the corresponding projections as 3D priors. While this improves 3D consistency, performance remains sensitive to inaccuracies in the projected point clouds. IDCNet~\cite{idcnet} jointly models depth and RGB spaces to inject more precise 3D information, but it requires large-scale depth video supervision, which limits its scalability. In our work, we investigate how confidence learning can be leveraged to better utilize noisy 3D priors.

Furthermore, recent studies explore utilizing video diffusion backbones as generative priors for Gaussian Splatting, thereby enhancing the generative capability of feedforward approaches~\cite{splatflow,wonderland,videorfsplat,generativegaussiansplatting}. However, adapting information from a video diffusion backbone to a Gaussian parameter decoder remains challenging since purely pixel-space supervision without explicit 3D constraints often leads to suboptimal results.}

{\parskip=2pt
\noindent\textbf{Confidence Learning in 3D Reconstruction}: 
An emerging direction in 3D reconstruction is the modeling of uncertainty. Klasson~\textit{et~al.}~\cite{klasson2024sourcesuncertainty3dscene} analyze the origins and types of uncertainty in GS~\cite{gs} and NeRF~\cite{nerf}. Several works improve reconstruction quality by introducing uncertainty estimation in Gaussian parameters~\cite{modelinguncertaintygaussiansplatting,han2025viewdependentuncertaintyestimation3d,wilson2024modelinguncertainty3dgaussian}, uncertainty-aware regularization~\cite{3dgsenhancer,kim20244dgaussiansplattingwild}, or modeling an uncertainty field~\cite{magicmomentsdifferentiableuncertainty,tan2025uncertaintyawarenormalguidedgaussiansplatting}. For diffusion-based methods~\cite{uncertaintyawarediffusionguidedrefinement,liu2024difflow3drobustuncertaintyawarescene,wang2024rigirectifyingimageto3dgeneration,wang2025lightweight}, uncertainty-aware losses are typically employed to balance training, assigning higher weights to regions with greater confidence.
In our work, we explore confidence learning with 3D foundation models, where confidence maps typically serve as auxiliary supervision during pretraining~\cite{vggt,dust3r,mast3r} or as state update rules~\cite{ttt3r}. We introduce a novel mechanism that assigns weights as confidence values derived from VGGT during noise initialization. Furthermore, inspired by the Kalman filter’s predict–update scheme for handling uncertainty, we design a module that fuses features with noisy measurements via learned residual corrections, enabling adaptive refinement of the representation.}

\begin{figure*}[pt]
    \vspace{5pt}
    \centering
    \includegraphics[width=0.95\linewidth]{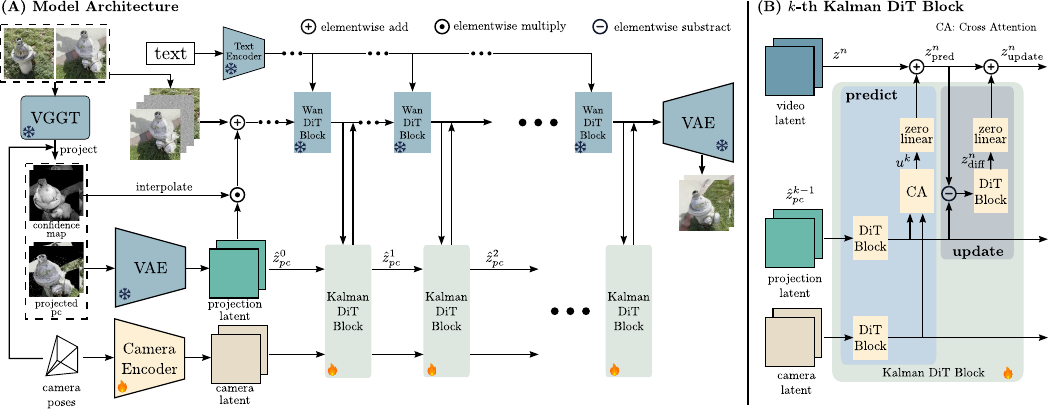}
      \caption{The \textbf{ConfCtrl} framework. (A)~\textit{Model Architecture.} Our framework builds upon the Wan2.1-Interpolation model~\cite{wan}. We introduce two key designs to enhance precise camera-following capability: (1)~noise initialization using a confidence-weighted projected point cloud latent, and (2)~predict-update camera conditioning mechanism that balance the predicted state with the projected point cloud as a noisy measurement. (B)~\textit{Kalman DiT Block.} It consists of two submodules: predict and update.}
    \vspace{-10pt}
    \label{fig:method}
\end{figure*}
\section{Preliminary}
\label{sec:preliminary}

{\parskip=2pt
\noindent\textbf{Adaptive Correction for Uncertainty in Kalman Filter}: The Kalman filter is a recursive algorithm that manages uncertainty in state estimation through a \emph{predict--update} loop, explicitly weighting predictions and measurements according to their respective confidences. First, the state $\hat{x}$ is projected forward in the prediction step:
\begin{equation}
\hat{x}_{\text{pred}} = f(\hat{x}, u),
\label{eqn:kalman1}
\end{equation}
where $u$ denotes the control input. The state is then corrected by integrating a new measurement $y$ in the update step:
\begin{equation}
\hat{x}_{\text{update}} = \hat{x}_{\text{pred}} + \Delta(\hat{x}_{\text{pred}}, y),
\label{eqn:kalman2}
\end{equation}
where $\Delta$ denotes the Kalman gain–weighted residual between the prediction and the measurement. This iterative process fuses predictions with noisy measurements to improve state estimation.

Motivated by the predict--update principle, we introduce a \emph{predict--update} structure as a per-block feature refinement module within our conditional control architecture (Sec.~\ref{sec:kalman}, Fig.~\ref{fig:method}(B)}). Unlike classical Kalman filters, it does not perform recursive Bayesian state estimation across diffusion steps, as diffusion~\cite{ddpm} violates linear system assumptions. Instead, it serves as an architectural mechanism to integrate control signals and geometric observations. Each block contains a \textit{prediction} submodule that generates features conditioned on the control input $u$ (e.g., target camera parameters), and an \textit{update} submodule that refines them using projected 3D point clouds as noisy measurements. This refinement improves geometric consistency and adherence to the target camera pose.

{\parskip=2pt
\noindent\textbf{Video Interpolation Conditioned on the First and Last Frames}: 
Video interpolation synthesizes intermediate frames between start and end frames while preserving temporal and spatial consistency. Given $\mathbf{I}_0$ and $\mathbf{I}_T$, the goal is to generate $\{\mathbf{I}_t\}_{t=1}^{T-1}$ forming a smooth transition. Modern methods~\cite{wan} learn motion and appearance transformations conditioned on the two endpoints, providing strong priors for camera control and view synthesis (Figure~\ref{fig:teaser}(A)).

\section{Method}
Our framework (Fig.~\ref{fig:method}) builds on a pretrained video interpolation model. To handle uncertainty in 3D priors, we introduce two components: a confidence-aware point cloud latent for noise initialization (Sec.~\ref{sec:noise_initialization}) and a predict–update conditioning mechanism for robust camera integration (Sec.~\ref{sec:kalman}). The model is trained with a rectified flow objective (Sec.~\ref{sec:training_loss}).%By leveraging the pretrained knowledge from video interpolation, our model efficiently adapts to novel view synthesis with minimal additional training, achieving both geometric precision and visual fidelity.

\subsection{Initialization Beyond Pure Noise}
\label{sec:noise_initialization}
Existing methods typically employ a ControlNet-like~\cite{controlnet} mechanism to condition the projected point cloud for camera control~\cite{gen3c, uni3c}. The projected point cloud is obtained from large-scale 3D foundation models~\cite{dust3r,mast3r,vggt,depthanything} and serves as a geometric prior. While these approaches perform well when the estimated point cloud is sufficiently accurate, their performance often degrades with sparse input views under large viewpoint changes.

In contrast, we propose to integrate this 3D prior into the \textit{initial distribution} for the rectified flow model, rather than initializing from pure Gaussian noise. To address inaccuracies in the point cloud, we introduce a \textit{point-wise confidence map} from VGGT that quantifies the reliability of each estimated point. This confidence map is projected into the target camera space to obtain a \textit{frame-wise confidence} that reflects the uncertainty for each target view.
Formally, given the projected point cloud latent $\hat{{z}}_{pc}^{0}$ and its corresponding confidence weights $\mathbf{w}$, we compute the confidence-aware initialization as:
\begin{equation}
    {z}_0 = \lambda_{1} \cdot (\mathbf{w} \odot \hat{{z}}_{pc}^{0}) +  \lambda_{2} \cdot \boldsymbol{\epsilon},
\label{eqn:noise_init}
\end{equation}
where $\odot$ denotes element-wise multiplication, $\boldsymbol{\epsilon} \sim \mathcal{N}(0, I)$ represents Gaussian noise, and $\lambda_1,\lambda_2 \in [0,1]$ control the balance between the confidence-aware latent and random noise. In our experiment, we select $\lambda_1=1,\lambda_2=1$. To ensure compatibility in temporal channel and spatial resolution, we apply linear interpolation to resize the projected point cloud latent before computing the weighted sum. This confidence-guided initialization provides a more reliable initial distribution, improving adaptation from video interpolation to novel view synthesis.

\subsection{Predict-Update Camera Conditioning}
\label{sec:kalman}
Existing methods for camera conditioning typically combine projected point clouds as a 3D geometric prior with camera embeddings (e.g., Plücker coordinates) to ensure pose controllability~\cite{uni3c,gen3c}. A common practice is to simply concatenate these two representations, which does not explicitly account for the inherent uncertainty and noise (e.g., occlusion, warping) present in projected point clouds.

To address this, we propose a \emph{predict–update} architecture that adaptively fuses geometric priors with camera control signals using learned residual corrections. Motivated by the predict--update paradigm in Kalman Filter (Sec.~\ref{sec:preliminary}), the process operates within each Kalman DiT block (see Figure~\ref{fig:method} (B)) and consists of two submodules: \textit{prediction} and \textit{update}. The prediction stage generates an initial feature conditioned solely on the target camera pose, while the update stage refines it by integrating the predicted feature with the noisy projected point cloud, effectively weighting the geometric observation according to its learned reliability.

\noindent\textbf{Prediction.} For the $k$-th Kalman DiT block, we treat the target camera pose as a control input $u^k$, which guides the video latent $z^{n}$ from the $n$-th Wan DiT block toward the desired viewpoint. The predicted latent $z^{n}_{\text{pred}}$ is defined as:
\begin{equation}
    z^{n}_{\text{pred}} =z^{n}+\operatorname{zero-linear}(u^{k})\triangleq f(z^{n}, u^{k}),
\end{equation}
analogous to the state transition in Kalman filtering (Eq.~\ref{eqn:kalman1}). The control input $u^{k}$ is implemented via a cross-attention mechanism between a camera latent (encoding the target camera Plücker embeddings) and the projection latent of the point cloud (which encodes the 3D prior and known camera poses). The resulting feature is then added back into $z^{n}$ through a zero-initialized linear layer, ensuring stable residual integration while preserving the original latent information. This submodule produces a predicted feature that combines the geometric prior with the intended camera motion. Alternative designs for this stage are evaluated in Table~\ref{tab:ablation}(f-g).

\noindent\textbf{Update.} In the \textit{update} submodule of the \( k \)-th Kalman DiT block, the projected point cloud latent \( \hat{z}_{pc}^{k-1} \) (i.e., \( \hat{z}_{pc}^{0} \) after passing through \( k-1 \) Kalman DiT blocks) is treated as a noisy measurement. To update the predicted latent $z^{n}_{\text{pred}}$ with  measurement $\hat{z}_{pc}^{k-1}$, we learn a correction residual
\begin{equation}
    z^{n}_{\text{diff}} = \operatorname{Diff}(z^{n}_{\text{pred}} - \hat{z}_{pc}^{k-1})
\end{equation}
using additional DiT blocks~\cite{dit}. These blocks, connected via  zero-initialized linear layers, output the residual based on the discrepancy between the prediction and the measurement. The final refined latent is then
\begin{equation}
    z^{n}_{\text{update}} =z^{n}_{\text{pred}} + \operatorname{zero-linear}(z^{n}_{\text{diff}}) \triangleq z^{n}_{\text{pred}} + \Delta(z^{n}_{\text{pred}},\hat{z}_{pc}^{k-1})
\end{equation}
analogous to the measurement update in Kalman Filter (Eq.~\ref{eqn:kalman2}). This residual-based fusion modulates the influence of geometric observations through the learned correction residual, attenuating inconsistent measurements and improving geometric consistency and camera controllability.

\subsection{Training Objective}
\label{sec:training_loss}
We adopt the standard \textit{rectified flow loss}~\cite{rf,flowmatching} as our training objective, but replace the initial distribution with our proposed \textit{confidence-aware latent projection}. This modification provides a more stable latent initialization to follow camera trajectories. Formally, given a latent variable $z_t$ at time $t$, we define its evolution under the rectified flow as
\begin{equation}
z_t = (1 - t) \, z_0 + t \, z_1,
\end{equation}
where $z_0$ is sampled from the confidence-aware latent prior (Eq.~\ref{eqn:noise_init}), and $z_1$ represents the target latent derived from the ground-truth video representation.  
The model learns a velocity field $v_\theta(z_t, t)$ that predicts the instantaneous flow between $z_0$ and $z_1$. The rectified flow loss is then given by
\begin{equation}
\mathcal{L}_{\text{RF}} = 
\| v_\theta(z_t, t) - (z_1 - z_0) \|_2^2,
\end{equation}
where the difference $z_{\text{target}}=(z_1 - z_0)$ acts as the supervision signal for the flow direction.  

To improve training stability and maintain fine-grained structural consistency when adapting video interpolation for novel view synthesis under noisy or jerky camera poses, we introduce a \textit{latent gradient regularization} term that enforces alignment of spatial gradients in the latent space:
\begin{equation}
\mathcal{L}_{\text{grad}} =
\| \nabla_x v_{\theta} - \nabla_x z_{\text{target}} \|_1 
+ \| \nabla_y v_{\theta} - \nabla_y z_{\text{target}} \|_1.
\end{equation}
The overall training objective is thus defined as
\begin{align}
\mathcal{L}_{\text{total}} &=
\mathcal{L}_{\text{RF}} + \lambda_{\text{grad}}\mathcal{L}_{\text{grad}},
\label{eqn:loss}
\end{align}
with $\lambda_{\text{grad}} = 0.05$ in our experiments. This regularization encourages the preservation of high-frequency details and enforces local spatial consistency, slightly reducing noise and flicker artifacts under rapid viewpoint changes while improving the model’s ability to accurately follow the camera trajectory (Table~\ref{tab:ablation}(e)).

\section{Experimental Results}
\label{sec:experiment}

\subsection{Setup}
\textbf{Dataset.} We train and evaluate our method on three datasets respectively: CO3D-Hydrant~\cite{co3d}, CO3D-Teddybear~\cite{co3d}, and DL3DV~\cite{dl3dv}, all rendered at a resolution of $256\times256$ pixels due to limited computational resources. Specifically, we use only 250, 450, and 200 sequences from each dataset for training, and the remaining sequences for testing.

\noindent\textbf{Metric.} We report the PSNR, LPIPS, and SSIM metrics to assess the quality of novel view synthesis, and \textit{translation error} (m) and \textit{rotation error} (radian) to evaluate camera controllability. The predicted camera poses are obtained using VGGT~\cite{vggt}, which estimates the camera trajectory from the generated video with respect to the first input image. The translation error is computed as the mean Euclidean distance between the predicted and ground-truth camera positions:
\begin{equation}
E_t = \frac{1}{N}\sum_{i=1}^{N} \| \mathbf{t}_i^{\text{pred}} - \mathbf{t}_i^{\text{gt}} \|_2,
\label{eqn:trans_error}
\end{equation}

where $\mathbf{t}_i^{\text{pred}}$ and $\mathbf{t}_i^{\text{gt}}$ denote the predicted and ground-truth camera translations, respectively. The rotation error measures the angular difference between predicted and ground-truth camera orientations:
\begin{equation}
E_r = \frac{1}{N}\sum_{i=1}^{N} \arccos\!\left(\frac{\mathrm{trace}(\mathbf{R}_i^{\text{pred}}\mathbf{R}_i^{\text{gt}^\top}) - 1}{2}\right),
\label{eqn:rot_error}
\end{equation}
where $\mathbf{R}_i^{\text{pred}}$ and $\mathbf{R}_i^{\text{gt}}$ represent the predicted and ground-truth rotation matrices. %These metrics jointly evaluate both translational and rotational accuracy of the predicted camera motions.

\noindent\textbf{Implementation Details.} Our implementation is built upon the 1.3B Wan-2.1-InP model~\cite{wan}. We insert 10 Kalman DiT blocks uniformly into the architecture to incorporate camera conditioning. For video diffusion, we use sequences of 33 frames. The model is trained for 20k steps with a learning rate of $10^{-5}$ until convergence. Due to limited computational resources and leveraging the strong prior of the interpolation model, we adopt a batch size of 1 and train on a single GPU, with a memory footprint of approximately 40 GB. Training for 20k steps takes about 24 hours on a single GPU, while inference with 50 diffusion steps takes approximately 10 seconds per scene.

\begin{table*}[hbpt]
\centering
\caption{Comparison of methods on three datasets. \textbf{Best} and \underline{second best} results are highlighted. For input camera condition, Cam denotes camera parameters, and PC denotes projected point cloud.}
\begin{adjustbox}{max width=\linewidth}
\begin{tabular}{l|cc|lllll|lllll|lllll}
\toprule
\multirow{2}{*}{\textbf{Method}} 
& \multirow{2}{*}{\textbf{Cam}} 
& \multirow{2}{*}{\textbf{PC}} 
& \multicolumn{5}{c|}{\textbf{CO3D-Hydrant}} 
& \multicolumn{5}{c|}{\textbf{CO3D-Teddybear}} 
& \multicolumn{5}{c}{\textbf{DL3DV}} \\
\cmidrule{4-18}
&  &  
& \textbf{LPIPS}~$\downarrow$ & \textbf{PSNR}~$\uparrow$ & \textbf{SSIM}~$\uparrow$ & $\mathbf{E_t}~\downarrow$ & $\mathbf{E_r}~\downarrow$ 
& \textbf{LPIPS}~$\downarrow$ & \textbf{PSNR}~$\uparrow$ & \textbf{SSIM}~$\uparrow$ & $\mathbf{E_t}~\downarrow$ & $\mathbf{E_r}~\downarrow$ 
& \textbf{LPIPS}~$\downarrow$ & \textbf{PSNR}~$\uparrow$ & \textbf{SSIM}~$\uparrow$ & $\mathbf{E_t}~\downarrow$ & $\mathbf{E_r}~\downarrow$ \\
\midrule
NopoSplat~\cite{noposplat} & \checkmark &  & 0.561 & 14.23 & 0.246 & {0.212} & {0.151} & 0.553 & 14.95 & 0.418 & 0.397 & 0.321 & 0.475 & 14.58 & 0.377 & 0.505 & 0.377 \\
DepthSplat~\cite{depthsplat} & \checkmark &  & 0.537 & 14.22 & 0.254 & 0.464 & 0.373 & 0.503 & 14.97 & 0.373 & 0.438 & 0.369 & 0.478 & 14.05 & 0.352 & 0.513 & 0.423 \\
MvSplat~\cite{mvsplat} &\checkmark  &  & 0.710 & 12.94 & 0.228 & 1.372 & 1.110 & 0.629 & 14.96 & 0.398 & 0.889 & 0.729 & 0.530 & 13.69 & 0.373 & 0.532 & 0.401 \\
PixelSplat~\cite{pixelsplat} &\checkmark  &  & 0.537 & 14.81 & 0.317 & 0.266 & 0.209 & 0.560 & 15.88 & \underline{0.462} & 0.637 & 0.528 & 0.476 & 15.56 & \underline{0.424} & 0.401 & 0.322 \\
AnySplat~\cite{anysplat} &\checkmark  &  & 0.368 & 15.36 & \underline{0.322} & 0.201 & 0.193 & 0.395 & 16.44 & 0.454 & 0.345 & 0.238 & 0.303 & 16.13 & 0.408 & 0.221 & 0.187 \\
LVSM~\cite{lvsm} & \checkmark &  & 0.435 & 14.25 & 0.288 & 0.282 & 0.254 & 0.415 & 15.26 & 0.407 & 0.395 & 0.386 & 0.326 & 15.73 & 0.374 & 0.279 & 0.245 \\
CameraCtrl~\cite{cameractrl} &\checkmark  &  & 0.415 & 14.51 & 0.285 & 0.260 & 0.194 & 0.461 & 14.95 & 0.394 & 0.716 & 0.565 & 0.343 & 15.25 & 0.347 & 0.276 & 0.222 \\
ViewCrafter~\cite{viewcrafter} &  & \checkmark & 0.413 & 14.61 & 0.291 & 0.249 & 0.181 & 0.408 & 15.94 & 0.419 & 0.361 & 0.271 & 0.324 & 15.40 & 0.370 & 0.264 & 0.201 \\
SeVA~\cite{seva} & \checkmark &  & 0.408 & 14.56 & 0.232 & 0.302 & 0.284 & 0.409 & 16.04  & 0.431 & 0.343 & 0.302 & 0.319 & 15.80 & 0.373 & 0.253 & 0.198 \\
Gen3R~\cite{gen3r} & \checkmark &  & \underline{0.341} & 15.32 & 0.321 & \underline{0.189} & \underline{0.144} & \underline{0.368} & \underline{16.92} & 0.452 & \underline{0.234} & \underline{0.159} & 0.296 & \underline{16.34} & 0.391 & 0.225 & 0.160 \\
Uni3C~\cite{uni3c} & \checkmark & \checkmark & 0.350 & \underline{15.41} & \underline{0.322} & 0.219 & 0.167 & 0.382 & 16.45 & 0.438 & 0.271 & 0.198 & \underline{0.306} & 15.99 & 0.395 & \underline{0.213} & \underline{0.159} \\
ConfCtrl (Ours) & \checkmark & \checkmark & \textbf{0.339} & \textbf{15.54} & \textbf{0.328} & \textbf{0.143} & \textbf{0.103} & \textbf{0.354} & \textbf{17.27} & \textbf{0.480} & \textbf{0.210} & \textbf{0.155} & \textbf{0.287} & \textbf{16.49} & \textbf{0.427} & \textbf{0.195} & \textbf{0.149} \\
\bottomrule
\end{tabular}
\end{adjustbox}

\label{tab:results}
\end{table*}
\begin{table*}[h]
\centering
\caption{Cross-dataset generalization results on CO3D, DL3DV, and GraspNet benchmarks. 
\textbf{Best} and \underline{second best} results are highlighted.}

\label{tab:cross_dataset}
\footnotesize
\setlength\tabcolsep{2.5pt}
\begin{tabular}{l|lllll|lllll|lllll}
\toprule
\multirow{2}{*}{\textbf{Method}} & \multicolumn{5}{|c}{\textbf{CO3D-Hydrant $\rightarrow$ CO3D-Teddybear}} &
\multicolumn{5}{|c|}{\textbf{CO3D-Hydrant $\rightarrow$ DL3DV}} &
\multicolumn{5}{c}{\textbf{DL3DV $\rightarrow$ GraspNet}} \\
\cmidrule{2-16}
& \multicolumn{1}{|c}{\textbf{LPIPS}~$\downarrow$} & \multicolumn{1}{c}{\textbf{PSNR~$\uparrow$}} &
\multicolumn{1}{c}{\textbf{SSIM~$\uparrow$}} & \multicolumn{1}{c}{$\mathbf{E_t}~\downarrow$} &
\multicolumn{1}{c}{$\mathbf{E_r}~\downarrow$} & \multicolumn{1}{|c}{\textbf{LPIPS}~$\downarrow$} &
\multicolumn{1}{c}{\textbf{PSNR}~$\uparrow$} & \multicolumn{1}{c}{\textbf{SSIM}~$\uparrow$} &
\multicolumn{1}{c}{$\mathbf{E_t}~\downarrow$} & \multicolumn{1}{c}{$\mathbf{E_r}~\downarrow$} &
\multicolumn{1}{|c}{\textbf{LPIPS}~$\downarrow$} & \multicolumn{1}{c}{\textbf{PSNR}~$\uparrow$} &
\multicolumn{1}{c}{\textbf{SSIM}~$\uparrow$} & \multicolumn{1}{c}{$\mathbf{E_t}~\downarrow$} &
\multicolumn{1}{c}{$\mathbf{E_r}~\downarrow$} \\
\midrule
NopoSplat~\cite{noposplat} & 0.656 & 14.13 & 0.293 & 1.125 & 0.949 & 0.687 & 10.01 & 0.172 & 1.099 & 0.878 & 0.659 & 10.87 & 0.242 & 0.222 & 0.157 \\
DepthSplat~\cite{depthsplat} & 0.675 & 12.48 & 0.290 & 1.218 & 1.032 & 0.659 & 10.65 & 0.201 & 1.079 & 0.854 & 0.621 & 10.02 & 0.189 & 0.202 & 0.143 \\
MvSplat~\cite{mvsplat} & 0.724 & 7.74 & 0.119 & 1.445 & 1.214 & 0.715 & 10.26 & 0.202 & 1.120 & 0.876 & 0.670 & 10.48 & 0.257 & 0.357 & 0.255 \\
PixelSplat~\cite{pixelsplat} & 0.641 & 13.93 & 0.402 & 0.691 & 0.550 & 0.643 & 12.82 & 0.293 & 0.701 & 0.529 & 0.576 & 14.42 & 0.405 & 0.219 & 0.156 \\
CameraCtrl~\cite{cameractrl} & 0.460 & 15.24 & 0.392 & 0.560 & 0.433 & 0.408 & 14.21 & 0.311 & 0.392 & 0.309 & 0.490 & 11.77 & 0.369 & 0.220 & 0.156 \\
ViewCrafter~\cite{viewcrafter} & \underline{0.423} & \underline{16.03} & \underline{0.408} & {0.359} & {0.261} & {0.352} & {15.29} & {0.354} & \underline{0.296} & \underline{0.234} & {0.272} & 14.37 & {0.476} & {0.180} & {0.127} \\
Uni3C~\cite{uni3c} & 0.439 & 15.26 & 0.405 & \underline{0.347} & \underline{0.256} & \underline{0.350} & \underline{15.30}& \underline{0.368} & 0.317 & 0.249 & \underline{0.238} & \underline{15.05} & \underline{0.499} & \underline{0.128} & \underline{0.091} \\
% \rowcolor{gray!10}
ConfCtrl (Ours) & \textbf{0.386} & \textbf{16.15} & \textbf{0.416} & \textbf{0.323} & \textbf{0.236} & \textbf{0.336} & \textbf{15.33} & \textbf{0.373} & \textbf{0.285} & \textbf{0.220} & \textbf{0.229} & \textbf{15.23} & \textbf{0.508} & \textbf{0.099} & \textbf{0.071} \\
\bottomrule
\end{tabular}
\end{table*}

% \begin{table}[htbp]

% \scriptsize

% \centering
% \resizebox{0.5\linewidth}{!}{
% \begin{tabular}{lcccc}
% \hline
%     & CameraCtrl & ViewCrafter & Uni3C  & Ours   \\ \hline
% FID & 127.82     & 129.32      & 124.53 & \textbf{109.08} \\
% FDDinov2 & 221.23     & 208.24      & 202.31 & \textbf{181.76} \\ 
% FVD & 146.26     & 148.02      & 140.98 & \textbf{123.86} \\ \hline
% \end{tabular}
% }
% \caption{Generative metrics comparison}
% \label{tab:rebuttal_fid}

% \end{table}

\begin{table}[hbpt]
\scriptsize
\centering
\begin{adjustbox}{max width=0.95\linewidth}
\begin{tabular}{c|ccc|ccc}
\toprule
            & \multicolumn{3}{c|}{CO3D-Hydrant} & \multicolumn{3}{c}{DL3DV} \\ \cmidrule{2-7} 
            & FID $\downarrow$       & FD-Dinov2 $\downarrow$    & FVD $\downarrow$      & FID $\downarrow$    & FD-Dinov2 $\downarrow$  & FVD $\downarrow$     \\ \midrule
CameraCtrl~\cite{cameractrl}  & 127.82    & 221.23     & 146.26   & 93.36 & 263.29   & 125.90 \\
ViewCrafter~\cite{viewcrafter} & 129.32    & 208.24     & 148.02   & 97.83 & 261.16   & 128.01 \\
Uni3C~\cite{uni3c}       & 124.53    & 202.31     & 140.98   & 89.21 & 233.07   & 118.64 \\
Gen3R~\cite{gen3r}       & 115.85    & 203.47     & 125.05   & 86.53 & 255.63   & \textbf{105.52} \\
Ours        & \textbf{109.08}    & \textbf{181.76}     & \textbf{123.86}   & \textbf{85.66} & \textbf{231.29}   & 113.46 \\ \bottomrule
\end{tabular}
\end{adjustbox}
\caption{Generative metrics comparison}
\label{tab:rebuttal_fid}
\end{table}

\begin{figure*}[t]
    \vspace{5pt}
    \centering
    \includegraphics[width=0.95\linewidth]{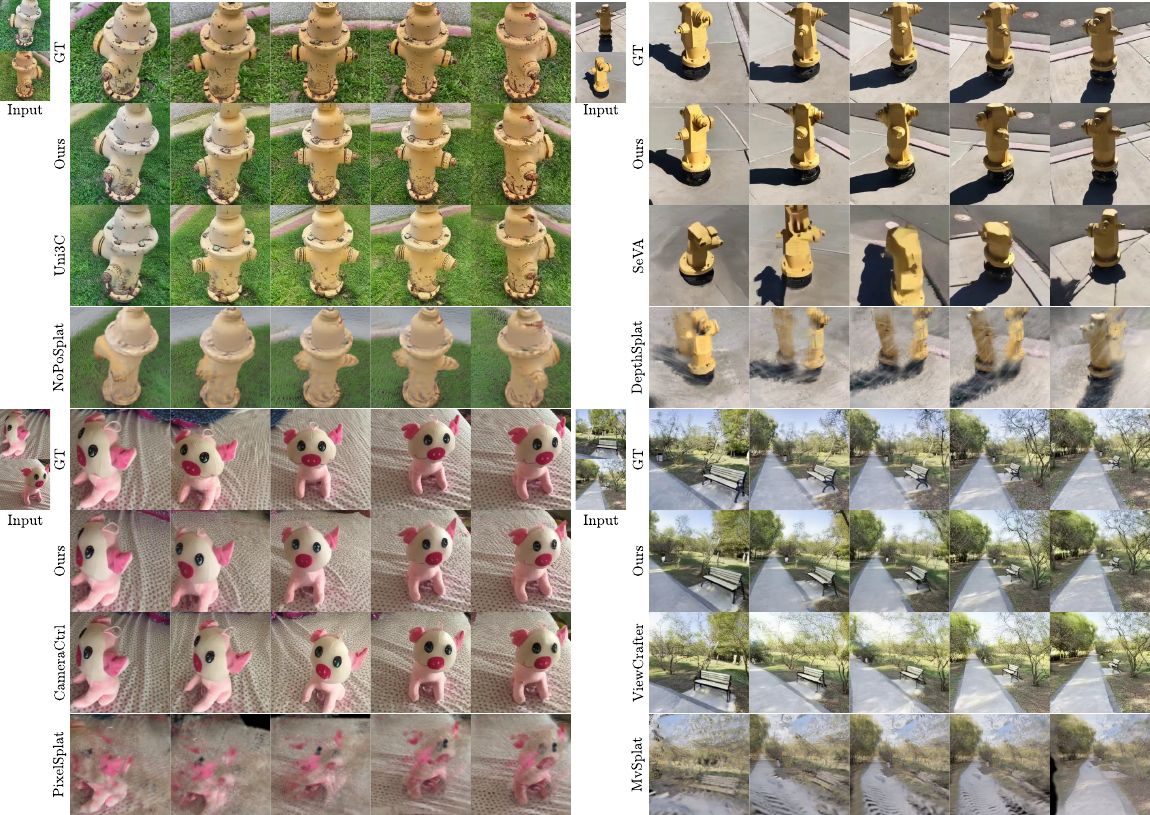}
    \caption{Qualitative comparison with baseline methods on the CO3D-Hydrant, CO3D-Teddybear, and DL3DV datasets. The two input images exhibit a large viewpoint change, demonstrating the robustness of our method under challenging view variations.}
    \vspace{-5pt}
    \label{fig:baseline}
\end{figure*}

\noindent\textbf{Baselines.} We compare our method with representative regression- and diffusion-based approaches trained on the same data. The regression baselines include PixelSplat~\cite{pixelsplat}, MvSplat~\cite{mvsplat}, DepthSplat~\cite{depthsplat}, NopoSplat~\cite{noposplat}, AnySplat~\cite{anysplat}, and LVSM~\cite{lvsm}; the diffusion baselines include CameraCtrl~\cite{cameractrl}, View-Crafter~\cite{viewcrafter}, and Uni3C~\cite{uni3c}. Diffusion baselines are retrained in a video diffusion setting conditioned on the first and last frames, using the same base model as ours for fairness. For Uni3C and ViewCrafter, we use the same VGGT-projected point clouds as input. We additionally compare with large-scale pretrained diffusion models, SeVA~\cite{seva} and Gen3R~\cite{gen3r}. More details are provided in supplementary. 

\subsection{Comparison with Existing Methods}
Quantitative results in Table~\ref{tab:results} demonstrate that our method consistently outperforms all baseline approaches across all evaluation metrics. The improvements in reconstruction scores highlight the robustness of our approach, even under sparse input views and large viewpoint variations. Moreover, the reduced translation and rotation errors indicate enhanced controllability and precise camera-pose alignment, confirming that our model not only generates visually accurate results but also adheres more faithfully to the specified geometric constraints. To further evaluate generative capability, we compare our method with diffusion-based approaches using additional metrics~\cite{fddinov2}: FID, FD-Dinov2, and FVD, as summarized in Table~\ref{tab:rebuttal_fid}. Following this evaluation protocol, our method outperforms the baselines in most cases, demonstrating its superiority over others.

As shown in Figure~\ref{fig:baseline}, our method further produces novel views with finer details and sharper visual quality compared to regression-based approaches, which often suffer from inaccurate Gaussian parameter estimation. In contrast to diffusion-based methods, our approach exhibits superior consistency with the specified target camera pose, achieving geometrically accurate and structurally coherent results across views.

\begin{figure*}[hbtp]
    \vspace{5pt}
    \centering
    \includegraphics[width=.95\linewidth]{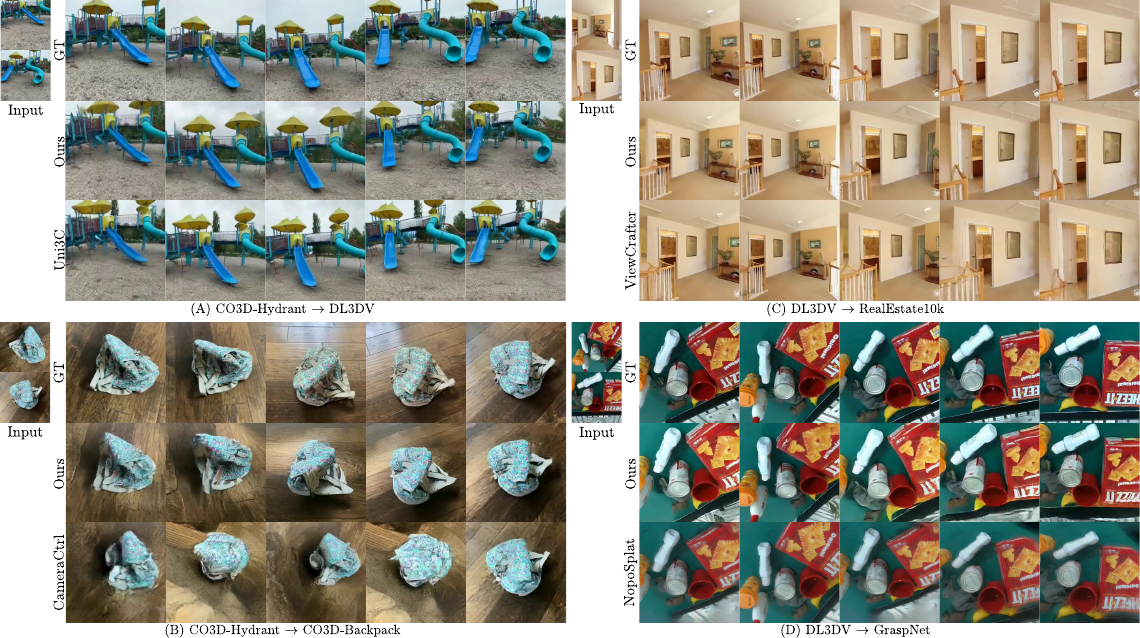}
    \caption{\textbf{Zero-shot results.} Qualitative comparison between our method and the ground truth on out-of-distribution datasets, demonstrating strong generalization ability.}
    \label{fig:zero_shot}
    \vspace{-5pt}
\end{figure*}

\subsection{Cross Dataset Performance}
Benefiting from the large-scale pretraining of video diffusion models, our method, along with other video-diffusion-based approaches, exhibits strong zero-shot generalization across diverse datasets compared to regression-based methods under large viewpoint changes. To rigorously evaluate cross-dataset performance, we conduct experiments on representative datasets including RealEstate10k~\cite{re10k}, GraspNet~\cite{graspnet}, CO3D~\cite{co3d}, and DL3DV~\cite{dl3dv}, which cover a wide range of scene types and motion distributions. Some quantitative comparisons are summarized in Table~\ref{tab:cross_dataset}. Our approach achieves the highest overall performance across all datasets, demonstrating superior robustness under significant domain shifts. These results indicate that the learned geometric priors and pose-conditioned generation framework effectively transfer to unseen scenarios without fine-tuning.

As illustrated in Figure~\ref{fig:zero_shot}, our method produces visually coherent and geometrically accurate novel views on out-of-distribution datasets. The generated results preserve fine scene details and maintain consistent spatial layouts across varying camera poses, demonstrating strong generalization of both appearance and geometry. These qualitative examples further confirm that our model can effectively handle diverse scene structures and camera trajectories without dataset-specific adaptation.
\begin{table}[tphb]
\centering
\caption{Ablation study of model components on CO3D-Hydrant dataset.}
\footnotesize
\setlength\tabcolsep{4.5pt}
\resizebox{0.98\columnwidth}{!}{%
\begin{tabular}{lccccc}
\toprule
 & \textbf{LPIPS}~$\downarrow$ & \textbf{PSNR}~$\uparrow$ & \textbf{SSIM}~$\uparrow$ & $\mathbf{E_t}~\downarrow$ & $\mathbf{E_r}~\downarrow$ \\
\midrule
(a) w/o noise init    & 0.347 & 15.44 & 0.326 & 0.174 & 0.128 \\
(b) w/o conf          & 0.366 & 14.75 & 0.296 & 0.164 & 0.117 \\
(c) w/o update        & 0.386 & 14.65 & 0.305 & 0.194 & 0.139 \\
(d) w/o both          & 0.362 & 14.85 & 0.304 & 0.213 & 0.155 \\ \midrule
(e) w/o $L_{grad}$    & 0.339 & 15.49 & 0.322 & 0.152 & 0.110 \\ \midrule
(f) camera pose only       & 0.370 & 14.90 & 0.312 & 0.192 & 0.141 \\
(g) camera pose $+$ proj.      & 0.363 & 15.06 & 0.313 & 0.174 & 0.124 \\ \midrule
(h) w/o interpolation & 0.491 & 13.45 & 0.193 & 0.457 & 0.362 \\
(i) param number             & 0.382 & 14.56 & 0.306 & 0.202 & 0.145 \\
\midrule
(j) VGGT $\rightarrow$ Dust3R   & 0.369 & 14.99 & 0.299 & 0.241 & 0.196 \\
(k) VGGT $\rightarrow$ Mast3R   & 0.356 & 15.07 & 0.314 & 0.207 & 0.161 \\
(l) VGGT $\rightarrow$ Pi3      & {0.333} & {15.56} & 0.326 & 0.146 & 0.107 \\
\midrule
(m) Kalman-DiT $\rightarrow$ MMDiT    & 0.366 & 15.01 & 0.314 & 0.194 & 0.140 \\ \midrule
Ours                  & 0.339 & 15.54 & 0.328 & 0.143 & 0.103 \\
\bottomrule
\end{tabular}%
}

\label{tab:ablation}

\end{table}
\begin{figure*}[t]

    \centering
    \includegraphics[width=\linewidth]{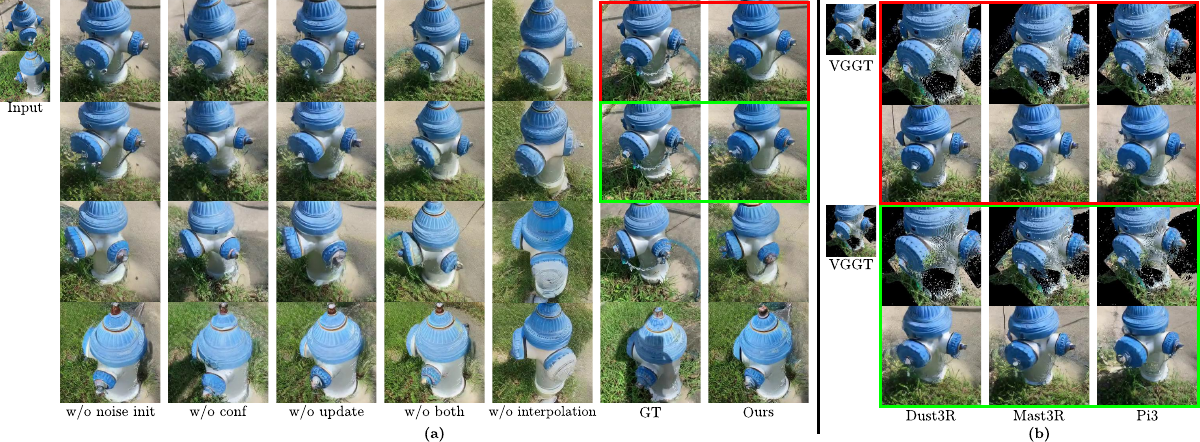}
    \caption{\textbf{Qualitative Ablation.} (a) Visual comparison of different model configurations showing the effect of each component. (b) Ablation across 3D foundation models. Our results and the ground truth (GT) are boxed in matching colors in (a).}

    \label{fig:ablation}
\end{figure*}
\subsection{Ablation Study}

\noindent\textbf{Noise Initialization.}
First, we remove the proposed confidence-weighted noise initialization and adopt the standard rectified flow starting from pure Gaussian noise. As shown in Table~\ref{tab:ablation}(a) and Fig.~\ref{fig:ablation}(a), both novel-view quality and camera control accuracy degrade significantly, with noticeable distortions and warping artifacts.

Second, we evaluate the role of the confidence map by assigning a uniform weight of 1.0. Results in Table~\ref{tab:ablation}(b) show that this naive setting further reduces image quality compared to (a), although it yields slightly better camera controllability.

\noindent\textbf{Kalman DiT Block.}
We remove the update submodule in each Kalman DiT block, making the model behave similarly to ControlNet~\cite{controlnet}, conditioned only on the camera pose and projected point cloud. Without this correction learning, the model struggles to balance the noisy 3D prior with the target camera pose, leading to noticeable degradation in both video quality and camera control accuracy(Fig.~\ref{fig:ablation}(a)), confirming that this update effectively refines the generation.

We further remove both the update submodule and the confidence-aware noise initialization. This variant yields the worst camera controllability among all settings, performing worse than (a), (b), and (c). The results indicate that Kalman DiT and confidence-weighted initialization are complementary, providing distinct benefits whose combination yields stronger performance.

\noindent\textbf{Gradient Regulation Loss.}
Removing the gradient loss as a regularization term in Eq.~\ref{eqn:loss} slightly degrades visual quality and camera control accuracy. While the gain is modest, this loss stabilizes training and improves spatiotemporal consistency.

\noindent\textbf{Camera Pose Condition in Prediction Submodule.}
In this experiment, we remove the cross-attention module that uses the projected point cloud latent in the prediction step of Kalman DiT, leaving the model conditioned only on the camera pose. The corresponding results are reported in Table~\ref{tab:ablation}(f). The observed performance drop confirms the importance of integrating both the camera pose and projected point cloud as complementary signals. While the simple addition of these features provides a minor improvement, it remains notably inferior to the cross-attention formulation (Table~\ref{tab:ablation} (g)).

\noindent\textbf{Interpolation Model.} We ablate the initialization choice by starting from a pretrained camera-guided video diffusion model instead of an interpolation model. As shown in Table~\ref{tab:ablation}(h) and Fig.~\ref{fig:ablation}(a), the substantial performance decline verifies that the interpolation model is crucial for preserving geometric consistency in novel view synthesis.

\noindent\textbf{Model Parameter Number.} Our Kalman DiT block comprises three DiT blocks. To verify that improvements stem from the architecture rather than parameter scaling, we build a deeper variant of design (c) with 15 layers to match our model's DiT count. Table~\ref{tab:ablation}(i) shows that naive scaling under the same training budget degrades performance, confirming the effectiveness of our design.

\noindent\textbf{3D Foundation Model.} To investigate the influence of different geometry foundation models, we replace VGGT in our methods with Dust3R~\cite{dust3r}, Mast3R~\cite{mast3r}, and Pi3~\cite{pi3}, respectively, providing projected point cloud and corresponding confidence map. Stronger geometry priors improve performance across all backbones while maintaining qualitative robustness (Table \ref{tab:ablation}(j-l),Fig.~\ref{fig:ablation}(b)), confirming the effectiveness of our design for precise camera control in video diffusion conditioned on noisy projected point clouds. 

\noindent\textbf{MMDiT.} MMDiT~\cite{mmdit} enables bidirectional information flow between pose and geometry signals. We replace our Kalman DiT block with an MMDiT block using the same inputs: the camera-pose latent and the projected point-cloud latent. As shown in Table~\ref{tab:ablation}(m), our method outperforms MMDiT, highlighting the advantage of our design. We attribute this gain to the update submodule, which refines latents using geometric observations alleviating uncertainty.

\section{Conclusion}
\label{sec:conclusion}

In this paper, we propose ConfCtrl, a framework built on a pretrained video interpolation model. ConfCtrl (1) uses a confidence-weighted point-cloud projection latent to initialize denoising, and (2) employs a predict–update camera conditioning mechanism to enable precise camera control and improve visual quality under noisy point-cloud conditions. Extensive experiments show that our approach outperforms both regression- and diffusion-based baselines across multiple benchmarks. On out-of-distribution datasets, it also demonstrates strong zero-shot performance.

\noindent\textbf{Limitations}: Despite these advantages, our model remains constrained by the limitations of the current VAE used in the video diffusion paradigm. Existing VAEs are primarily designed to generate temporally consistent, smooth content between adjacent frames, making them less suitable for novel view synthesis involving abrupt camera motion or large positional changes. Future work may explore optimizing the VAE architecture for this setting or eliminating the VAE~\cite{DiffRA,shi2025latent} from novel view generation.

\clearpage
{
    \small
    \bibliographystyle{ieeenat_fullname}
    \bibliography{main}
}

% WARNING: do not forget to delete the supplementary pages from your submission 
% \input{sec/X_suppl}

\end{document}